\documentclass{article}

     \usepackage[preprint,nonatbib]{neurips_2020}

\usepackage[dvipsnames]{xcolor}
\usepackage[utf8]{inputenc} %
\usepackage[T1]{fontenc}    %
\usepackage{hyperref}       %
\usepackage{url}            %
\usepackage{booktabs}       %
\usepackage{amsfonts}       %
\usepackage{nicefrac}       %
\usepackage{microtype}      %
\usepackage[ruled,vlined]{algorithm2e}

\usepackage[noadjust]{cite}
\usepackage{graphicx}
\usepackage[utf8]{inputenc} %
\usepackage[T1]{fontenc}    %
\usepackage{hyperref}       %
\usepackage{url}            %
\usepackage{booktabs}       %
\usepackage{amsfonts}       %
\usepackage{nicefrac}       %
\usepackage{microtype}      %
\usepackage{hyperref}
\usepackage{multirow, makecell}
\usepackage{longtable}

\SetKwFunction{Range}{range}
\SetKwInput{KwInput}{Input}
\SetKwInput{KwParams}{Parameters}

\newcommand{\xlsrp}{XLS-R}
\newcommand{\xlsrpb}[1]{\xlsrp{} {(#1B)}}

\newcommand{\xenglish}{X $\rightarrow$ English}

\def\textdataset{{Web-NTL}}

\begin{document}
\title{Google USM:\\
Scaling Automatic Speech Recognition\\
Beyond 100 Languages}

\author{Yu Zhang \And Wei Han\And James Qin\And Yongqiang Wang\And Ankur Bapna \And Zhehuai Chen \And 
Nanxin Chen \And Bo Li 
\And Vera Axelrod \And Gary Wang \And Zhong Meng  \And Ke Hu \And Andrew Rosenberg \And Rohit Prabhavalkar \And Daniel S. Park \And Parisa Haghani \And Jason Riesa  \And Ginger Perng   \And Hagen Soltau \And Trevor Strohman \And Bhuvana Ramabhadran \And Tara Sainath \And Pedro Moreno \And Chung-Cheng Chiu \And Johan Schalkwyk \And Françoise Beaufays
\And Yonghui Wu
\thanks{All authors are affiliated with Google Inc.}
\thanks{Contact author at \texttt{\scriptsize ngyuzh@google.com}.}}

\maketitle

\begin{abstract}
We introduce the Universal Speech Model (USM), a single large model that performs automatic speech recognition (ASR) across 100+ languages. This is achieved by pre-training the encoder of the model on a large unlabeled multilingual dataset of 12 million (M) hours spanning over 300 languages, and fine-tuning on a smaller labeled dataset.  We use multilingual pre-training with random-projection quantization and speech-text modality matching to achieve state-of-the-art performance on downstream multilingual ASR and speech-to-text translation tasks. We also demonstrate that despite using a labeled training set 1/7-th the size of that used for the Whisper model \cite{radford2022robust}, our model exhibits comparable or better performance on both in-domain and out-of-domain speech recognition tasks across many languages.
\end{abstract}

\section{Introduction}

Recent advances in self-supervised learning have ushered in a new era for speech recognition. Whereas previous works focused mostly on improving the quality of monolingual models for mainstream languages, recent studies have increasingly turned to “universal” models \cite{speechstew,bigssl,li2021scaling,radford2022robust}. These may take the form of a single model that performs well on multiple tasks \cite{speechstew,radford2022robust}, or one that covers multiple domains \cite{speechstew,bigssl}, or one that supports multiple languages \cite{li2022asr2k,radford2022robust}. In this work, we explore the frontiers of language expansion. Our long-term goal is to train a universal ASR model that covers all the spoken languages in the world. 

A fundamental challenge in scaling speech technologies to many languages is obtaining enough data to train high-quality models. With conventional supervised training approaches, audio data needs to be manually transcribed, which is lengthy and expensive, or collected from existing transcribed sources which are hard to find for tail languages. While transcribed speech may be scarce in many languages, untranscribed speech and text data are practically unlimited.
Recent developments in semi-supervised algorithms for speech recognition makes it possible to leverage such data for pre-training and produce high-quality speech models with a limited amount of transcribed data \cite{wav2vec2,bigssl}. 

Moreover, recent studies have shown that a single large model can utilize large data sets more effectively than smaller models \cite{li2021scaling,radford2022robust}. This all points to a promising direction where large amounts of unpaired multilingual speech and text data and smaller amounts of transcribed data can contribute to training a single large universal ASR model.

\subsection{Our approach}

We produce large ``Universal Speech Models" (USMs) through a training pipeline that utilizes three types of datasets:
\begin{itemize}
\item \textbf{Unpaired Audio:} 
 \begin{itemize}
 \item \textbf{YT-NTL-U}: A large unlabeled multilingual dataset consisting of 12M hours of YouTube-based audio covering over 300 languages.
 \item \textbf{Pub-U}: 429k hours of unlabeled speech in 51 languages based on public datasets.
 \end{itemize}
\item \textbf{Unpaired Text:} 
 \begin{itemize}
     \item \textbf{\textdataset}: A large multilingual text-only corpus with 28B sentences spanning over 1140 languages.
 \end{itemize}
\item \textbf{Paired ASR Data:} We utilize two corpora of paired audio-text data with O(10k) hours of audio for supervised training.
 \begin{itemize}
 \item \textbf{YT-SUP+}: 90k hours of labeled multilingual data covering 73 language and 100k hours of en-US pseudo-labeled data generated by noisy student training (NST) \cite{nst, nstasr} from YT-NTL-U.
 \item \textbf{Pub-S}: 10k hours of labeled multi-domain en-US public data and 10k labeled multilingual public data covering 102 languages.
 \end{itemize}
\end{itemize}

2B-parameter Conformer \cite{conformer} models are built using these datasets through the following steps:
\begin{enumerate}
\item \textbf{Unsupervised Pre-training:} BEST-RQ (\textbf{BE}RT-based \textbf{S}peech pre-\textbf{T}raining with \textbf{R}andom-projection \textbf{Q}uantizer) \cite{pmlr-v162-chiu22a} is used to pre-train the encoder of the model with YT-NTL-U.
\item \textbf{MOST} (\textbf{M}ulti-\textbf{O}bjective \textbf{S}upervised pre-\textbf{T}raining): The model can optionally be further prepared by a multi-objective supervised pre-training pipeline that utilizes all three kinds of datasets: YT-NTL-U, Pub-U, \textdataset~and Pub-S. Here, a weighted sum of the BEST-RQ masked language model loss \cite{bert}, along with the text-injection losses (including the supervised ASR loss and modality matching losses) \cite{chen2021injecting,chen2022maestro} is optimized during training.
\item \textbf{Supervised ASR Training}: We produce generic ASR models trained with connectionist temporal classification (CTC) \cite{ctc} and Listen, Attend, and Spell (LAS) \cite{las} tranducers for downstream tasks.
\end{enumerate}

Two types of models are produced through this pipeline---pre-trained models that can be fine-tuned on downstream tasks, and generic ASR models for which we assume no downstream fine-tuning occurs. The generic ASR models are trained with chunk-wise attention, which we introduce later in this report.
\begin{table}[h!]
  \caption{USM models prepared in this work. The generic ASR models are trained on a large "upstream" ASR corpus and not finetuned further, while the pre-trained models are fine-tuned on downstream tasks.}
  \label{t:usm-models}
  \centering
  \resizebox{0.95\columnwidth}{!}{%
  \begin{tabular}{lcccccc}
  \toprule
  Model & BEST-RQ & MOST & Model-Type & Decoder & Upstream & Chunk-wise \\
  & & & & & ASR Dataset & Attention \\
  \midrule
  \bf USM & \multirow{4}{*}{YT-NTL-U} & N & Pre-trained & Downstream Dependent & - & N \\
  \bf USM-M & & Y & Pre-trained &  Downstream Dependent &- & N \\
  \bf USM-LAS & & N & Generic ASR & LAS & YT-SUP+ & Y \\
  \bf USM-CTC & & N & Generic ASR & CTC & YT-SUP+ & Y \\
  \bottomrule
  \end{tabular}}
\end{table}

We denote the pre-trained models USM and USM-M, where the appendix \textbf{-M} indicates that MOST has been utilized for the preparation of the model. The USM and USM-M models can be further fine-tuned on the downstream task of choice with an appropriate transducer unit, which can be a CTC, LAS or RNN transducer (RNN-T) unit. We evaluate our USM models on two types of benchmarks:
\begin{itemize}
\item \textbf{Automatic Speech Recognition (ASR):} We use YouTube data to train USMs for YouTube (e.g., closed captions). We evaluate the USMs on two public benchmarks, SpeechStew \cite{speechstew} and FLEURS \cite{conneau2022fleurs}. We also report results on the long-form test set CORAAL \cite{kendall2021corpus} for which only the evaluation set is available.
\item \textbf{Automatic Speech Translation (AST):} We test AST performance on CoVoST 2 \cite{wang2020covost}.
\end{itemize}
As indicated in Table \ref{t:usm-models}, the generic ASR models are trained with YT-SUP+ and not fine-tuned on domain-specific datasets for downstream ASR tasks.  We, however, explore the possibility of attaching additional ``adapter" units \cite{he2022towards} to both generic and pre-trained ASR models and training adapter weights while keeping the rest of the model frozen.

\begin{figure}[h!]
\centering
\includegraphics[width=0.98\columnwidth]{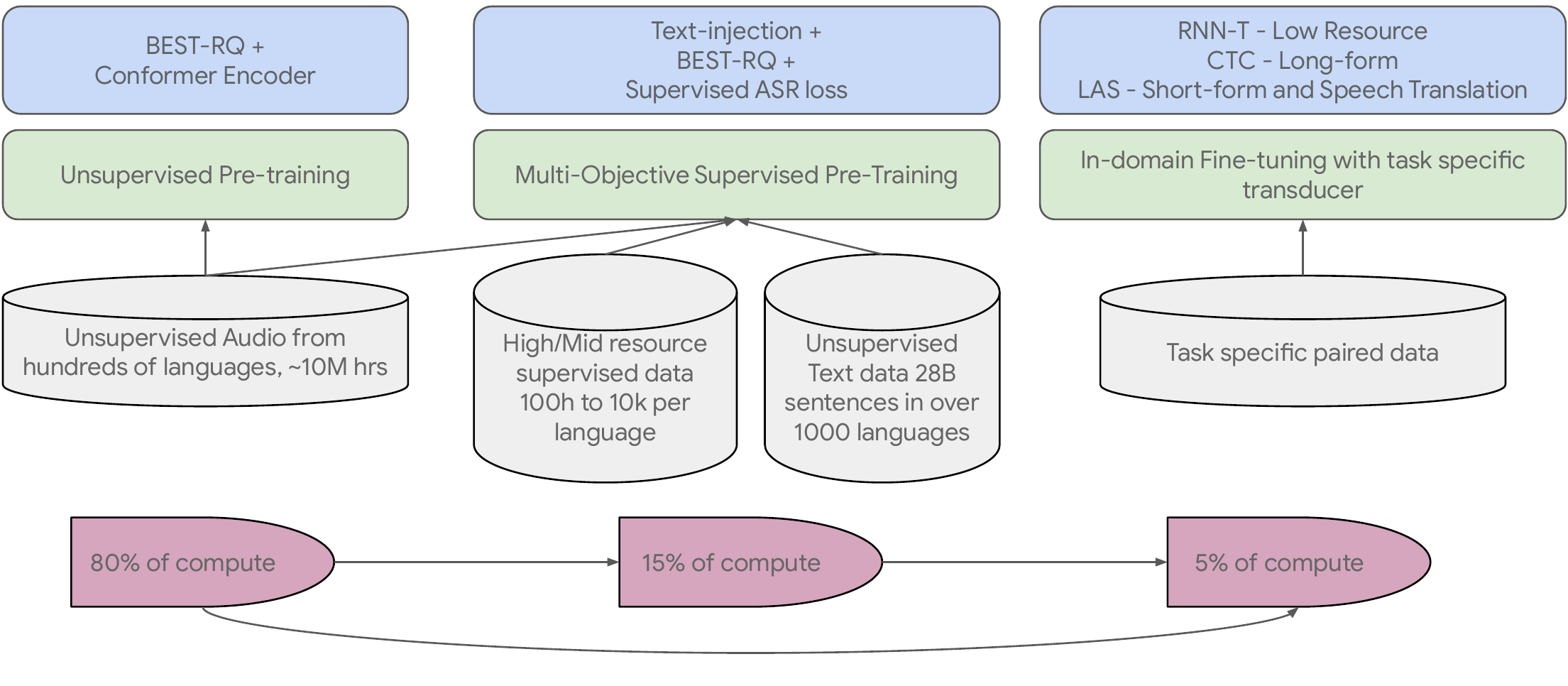}
\caption{An overview of our approach. Training is split into three stages. (i) The first stage trains a conformer backbone on a large unlabeled speech dataset, optimizing for the BEST-RQ objective. (ii) We continue training this speech representation learning model while optimizing for multiple objectives, the BEST-RQ objective on unlabeled speech, the modality matching, supervised ASR and duration modeling losses on paired speech and transcript data and the text reconstruction objective with an RNN-T decoder on unlabeled text. (iii) The third stage fine-tunes this pre-trained encoder on the ASR or AST tasks.}
\label{f:overview}
\end{figure}

The overall training pipeline of our models is summarized in Fig. \ref{f:overview}. In our design, once a large amount of compute is expended in the pre-training stages, the downstream application can be conveniently fine-tuned from a model trained from stage-1 or stage-2 with a task-specific transducer. Our experimental results demonstrate that this pipelined training framework enables us to build both generic multilingual ASR systems and domain specific models with state-of-the-art performance.

We next present the key findings of our research, provide an overall view of the report, and review related work.

\subsection{Key Findings}
\begin{figure*}[t!]
\centering
\includegraphics[width=0.97\textwidth]{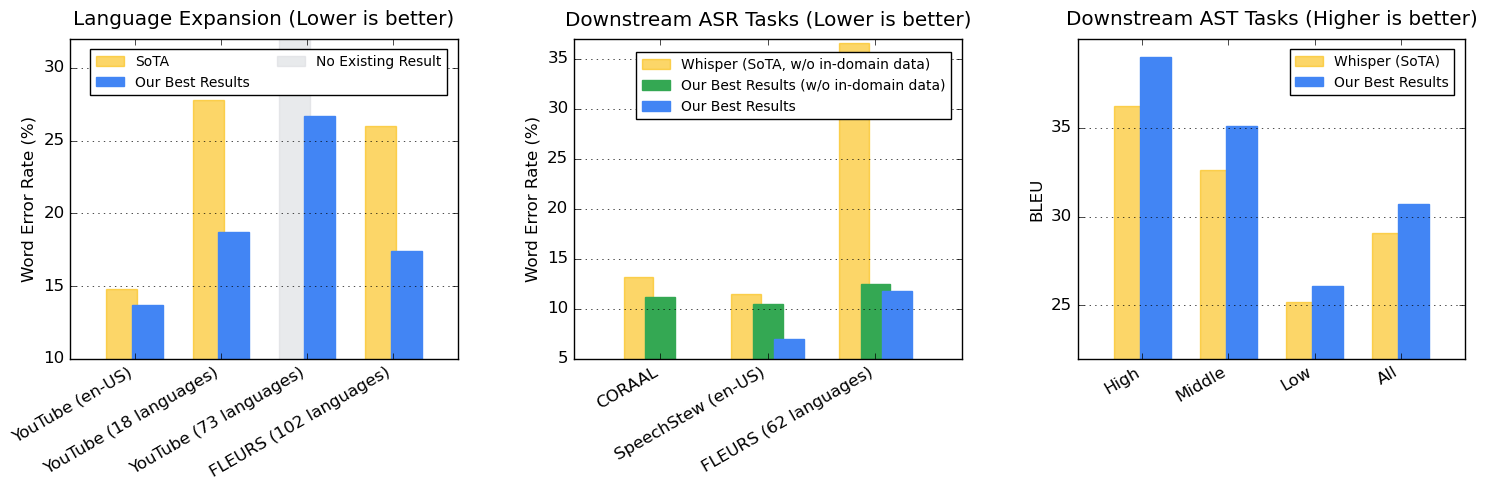}
\caption{\textbf{(Left)}$^\dagger$ WERs (\%) Our language expansion effort to support more languages on YouTube (73 languages) and extending to 100+ languages on the public dataset (FLEURS). Lower is better. To the best of our knowledge, no published model can successfully decode all 73 languages from our YouTube set, thus we only list our results. \textbf{(Middle)}$^\dagger$ Our results on ASR benchmarks, with or without in-domain data. Lower is better. \textbf{(Right)} SoTA results on public speech translation tasks. Results presented are presented as high/middle/low resources languages defined in \cite{bapna2022mslam}. Higher is better.}
\label{f:summary}
\end{figure*}

\textbf{SoTA results for downstream multilingual speech tasks:} Our USM models achieve state-of-the-art performance for multilingual ASR and AST for multiple datasets in multiple domains. This includes SpeechStew (mono-lingual ASR) \cite{speechstew}, CORAAL (African American Vernacular English (AAVE) ASR) \cite{kendall2021corpus}, FLEURS (multi-lingual ASR) \cite{conneau2022fleurs}, YT (multilingual long-form ASR), and CoVoST (AST from English to multiple languages). We depict our model's performance in the first panel of Fig. \ref{f:summary}. We also build an ASR model for YouTube captioning -- i.e., the transcription of speech in YouTube videos, that achieves < 30\% WER on 73 languages. With only 90k hours of supervised data, this model performs better than Whisper \cite{radford2022robust}, a strong general ASR system trained on more than 400k hours of transcribed data (we select 18 languages that Whisper can successfully decode with lower than 40\% WER). The second panel of Fig. \ref{f:summary} demonstrates that our YouTube captions model generalizes well to unseen domains.

\textbf{BEST-RQ is a scalable speech representation learner:} We find that BEST-RQ pre-training can effectively scale to the very large data regime with a 2B parameter Conformer-based backbone, comparing favorably against Wav2Vec 2.0 \cite{wav2vec2} and W2v-BERT \cite{chung2021w2v} in this setting. 

\textbf{MOST (BEST-RQ + text-injection) is a scalable speech and text representation learner:} We demonstrate that MOST is an effective method for utilizing large scale text data for improving quality on downstream speech tasks, as demonstrated by quality gains exhibited for the FLEURS and CoVoST 2 tasks. Fig.~\ref{f:summary} depicts USM's performance, establishing a new state-of-the-art on the FLEURS benchmark across 102 languages for ASR and on CoVoST 2 across 21 languages on AST.

\textbf{Representations from MOST (BEST-RQ + text-injection) can quickly adapt to new domains:}
We find that it is possible to obtain powerful downstream ASR/AST models by attaching and training light-weight residual adapter modules, which only add 2\% of additional parameters, while keeping the rest of the model frozen.

\textbf{Chunk-wise attention for robust long-form speech recognition: } We introduce chunk-wise attention, an effective, scalable method for extending the performance of ASR models trained on shorter utterances to very long speech inputs. We find that the USM-CTC/LAS models trained with chunk-wise attention is able to produce high-quality transcripts for very long utterances in the YouTube evaluation sets.

\subsection{Outline}

The outline of this report is as follows:

\textbf{Methods:} We review the architecture and the methods used in the paper. We provide brief summaries of the Conformer \cite{conformer}, BEST-RQ \cite{pmlr-v162-chiu22a}, text-injection \cite{chen2021injecting,chen2022maestro} used for MOST, and Noisy Student Training (NST) \cite{nst, nstasr}. We also introduce chunk-wise attention for scalable training on long utterances.

\textbf{Data:}  We describe the four types of datasets used to train our models: the unlabeled multilingual speech dataset YT-NTL-U, the  multilingual text corpus \textdataset, labeled datasets, and pseudo-labeled datasets.

\textbf{Key Results:} We present the performance of our USM models on downstream ASR and AST tasks. We demonstrate that USM establishes new states-of-the-art on several speech understanding benchmarks.

\textbf{Analysis and Ablations:} We present analysis of the effects of the key components of our work and compare their performance against existing methods.

\subsection{Related Work}
There is extensive literature on pre-training \cite{hsu2018extracting, chung2018speech2vec, oord2018representation, chung2019autoregressive, chorowski2019unsupervised, schneider2019wav2vec, baevski2019vqwav2vec, ling2019deep, baevski2019effectiveness, riviere2020unsupervised, kawakami2020learning, wav2vec2,babu2021xls,chen2021injecting} and self-training \cite{Zavaliagkos98utilizinguntranscribed,Lamel00lightlysupervised,Novotney2009,Thomas2013,li2019,kahn2019selftraining,synnaeve2019endtoend,parthasarathi2019,hsu2020selfsupervised,nstasr,xu2020iterative,chen21c_interspeech} for ASR.  Large speech models trained on large datasets have been studied previously in both monolingual \cite{bigssl} and multilingual contexts \cite{li2021scaling,radford2022robust}. Large multi-modal speech models have been explored in \cite{renduchintala2018multi,bapna2021slam,bapna2022mslam,chen2022maestro,thomas2022textogram,cheng2022mu,Zhang2022SpeechUTBS,Zhang2022SpeechLMES,Khurana2022SAMUXLSRSM,Zhou2022MMSpeechMM,sainath2023joist, zhong2023jeit}. Various unsupervised pre-training methods for speech models have been proposed and applied in \cite{wav2vec2,chung2021w2v,pmlr-v162-chiu22a}.

Our work is an extension of a host of recent research efforts \cite{bigssl, chen2022maestro, sainath2023joist,pmlr-v162-chiu22a,meng2023modular} that have studied semi-supervised learning for ASR in the context of deep-learning. Large speech models (> 1B) were first studied in \cite{bigssl}; we expand upon this approach to train multilingual speech models in this work. We improve the methods used in \cite{bigssl} by employing a more scalable self-supervised learning algorithm (BEST-RQ) and additionally applying multi-modal pre-training (text-injection) to prepare the models. We introduce an improvement to BEST-RQ \cite{pmlr-v162-chiu22a} by utilizing a multi-softmax loss. We also incorporate Multi-Objective Supervised Training (BEST-RQ with text-injection) to improve the quality of speech representations learnt during pre-training, by utilizing transcribed data and unlabeled text. Long-form ASR has been studied in \cite{chiu2019comparison,lu2021input,radford2022robust}; we propose chunk-wise attention as an alternative solution to chunk-based decoding.

In this paper, we propose a scalable self-supervised training framework for multilingual ASR which extends to hundreds of languages. In particular: 
\begin{itemize}
\item We demonstrate that USMs pre-trained on 300 languages can successfully adapt to both ASR and AST tasks in new languages with a small amount of supervised data.
\item We build a generic ASR model on 73 languages by fine-tuning pre-trained models on 90k hours of supervised data. We show that the generic ASR models can carry out inference efficiently on TPUs and can reliably transcribe hours-long audio on YouTube Caption ASR benchmarks.
\item We conduct a systematic study on the effects of pre-training, noisy student training, text injection, and model size for multilingual ASR.
\end{itemize}

\section{Methods}

\subsection{Model Architecture: Conformer}
\label{ss:conformer}

We use the convolution-augmented transformer \cite{conformer}, or Conformer, with relative attention \cite{dai2019transformer} as an encoder model. For downstream speech tasks such as ASR or AST, the features produced by the Conformer are either used as an input to a connectionist temporal classification (CTC) \cite{ctc}, RNN transducer (RNN-T) \cite{rnnt} or a Listen, Attend, and Spell (LAS) \cite{las} unit after additional projection. As will be discussed further, BEST-RQ pre-training is exclusively applied to the encoder, while other forms of training (e.g., T5 \cite{raffel2020exploring}) train the entire task network as a whole.

For our experiments, we consider two models with 600M and 2B parameters respectively. While the main results presented have been obtained using the 2B model, the 600M model is utilized for ablation studies and observing model scaling behavior. Some features of the models are listed in Table \ref{t:cparams}. 

\begin{table}[h!]
  \caption{Conformer model parameters.}
  \label{t:cparams}
  \centering
  \resizebox{0.8\columnwidth}{!}{%
  \begin{tabular}{lcccccc}
    \toprule
    Model & \# Params (B) & \# Layers & Dimension & Att. Heads & Conv. Kernel Size \\
    \midrule
    Conformer-0.6 & 0.6 & 24 & 1024 & 8 & 5\\
    Conformer-2B & 2.0 & 32 & 1536 & 16 & 5\\
    \bottomrule
  \end{tabular}}
\end{table}

\subsection{Pre-training: BEST-RQ}

\begin{figure*}[t!]
\centering
\includegraphics[width=0.8\textwidth]{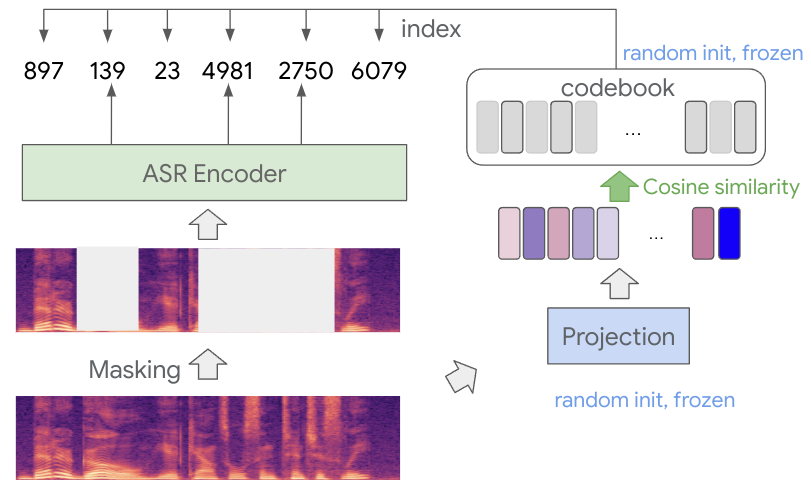}
\caption{BEST-RQ based pre-training with conformer encoder.}
\label{f:best-rq}
\end{figure*}

We select BEST-RQ \cite{pmlr-v162-chiu22a} as the method to pre-train our networks with speech audio. BEST-RQ provides a simple framework with a small number of hyperparameters for unsupervised training on large-scale unlabeled audio data. We discuss the comparative advantage of BEST-RQ against other pre-training methods in section \ref{sec:best-rq}.

BEST-RQ employs a BERT-style training task for the audio input that attempts to predict masked speech features. To make the task compatible with BERT-style training, the original speech features corresponding to the masked frames are quantized, and the task requires predicting the quantized label of these features. For a given number of quantization targets $c$, random ``codebook" vectors $v_0, \cdots, v_{c-1}$ are chosen in an embedding space. The discrete label of the speech feature is obtained by first projecting the feature into the embedding space by a randomly initialized, frozen projection matrix and then finding the closest codebook vector. The index of this codebook vector is identified as the label of the speech feature. Cosine similarity is used as the distance measure for determining the code. 

We note that while w2v-BERT \cite{chung2021w2v} pre-training has proven to be an effective method for unsupervised pre-training, it requires an additional quantization module which introduces more complexity. As we increase the model size and language coverage, the learnt codebook module proves costly to tune and can impede progress of model development. Meanwhile, the BEST-RQ algorithm does not require such a module, making it a more scalable method for pre-training.

\subsubsection{Multi-softmax}

Instead of utilizing a single codebook \cite{pmlr-v162-chiu22a}, we use multiple codebooks to improve BEST-RQ training in this study. More precisely, we use $N$ softmax layers to produce $N$ probability predictions from the output of the encoder to compare against $N$ independent quantization targets obtained from the masked speech features. We train the network with equal weights for each softmax layer. The use of multiple codebooks improves the stability and convergence of the model.

\subsection{Self-training: Noisy Student Training}

We utilize noisy student training (NST) \cite{nst, nstasr} to generate pseudo-labeled data to augment supervised training. This is done by first training a teacher model with augmentation on a supervised set, then using that teacher to generate transcripts for unlabeled audio data. A heuristic filtering method based on the ratio between the number of words and audio length is used to filter the pseudo-labeled data. The pseudo-labeled data is mixed with supervised data to train the student model.

\subsection{Chunk-wise Attention for Long-form ASR}
\label{ss:chunk-wise-attention}

\begin{figure}[h!]
\centering
\includegraphics[width=0.8\columnwidth]{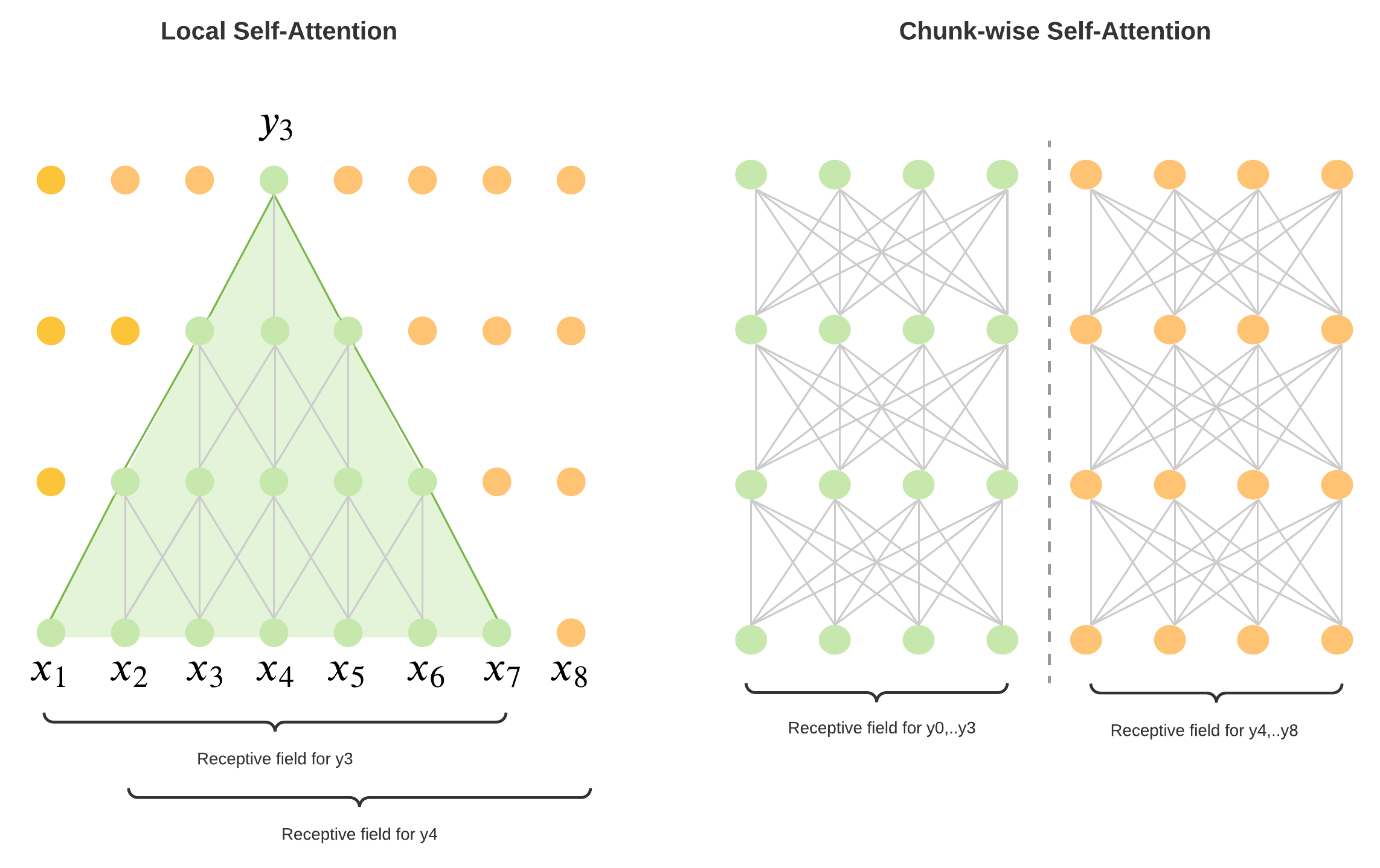}
\caption{Comparing receptive fields of two networks with 4 layers of local self attention and chunk-wise attention.}
\label{f:receptive_mismatch}
\end{figure}

In many real-world applications, ASR systems are required to transcribe minutes- or hours-long audio. This poses significant challenges to many end-to-end ASR systems, as these ASR systems are usually trained on much shorter segments, typically less than 30 seconds. For systems that use attention-based encoders, it is impractical to use global attention to attend to the entire audio. Local self attention, which only attends to the fixed length of left and right context, is thus widely used. For example, in BEST-RQ pre-training, only 128 left and 128 right context frames are used for local self attention. However, stacking many local self attention layers creates a significant receptive field mismatch between training and inference. The left figure in Fig. \ref{f:receptive_mismatch} illustrates this issue with a network consisting of 4 local self attention layers, each using only 1 left and 1 right context frames. Since the context is leaked in every layer, the receptive field width grows linearly with respect to the number of layers; for a big encoder like that of the Conformer-2B, this means that the receptive field width for the encoder output is longer than 327 seconds. During training, the model is trained with at most 30 seconds speech segments, while at inference time, when minutes or hours long audio is fed to the model, the encoder needs to process over 300 seconds of audio to produce one encoder output---a pattern it has never trained on. Our empirical observations demonstrate that, under this train-test mismatch, these models with deep architectures and high capacity suffer from high deletion errors. We henceforth refer to this problem as the ``long-form (performance) degradation" problem.

To solve this problem, we propose a simple modification to the attention mechanism; the attention is restricted to audio chunks. This is illustrated on the right side of Fig. \ref{f:receptive_mismatch}, in which 8 frames are divided into 2 chunks, and the attention is performed within each chunk. In this case, there is no context leaking in the attention layer, and thus the receptive field width is independent of the number of layers. In our experiments an 8-second chunk resulted in the best recognition quality vs. computational cost trade-off. 

It is worthwhile to note there are a few other works in the literature which also modify the attention pattern to deal with the long-form audio in ASR, e.g., \cite{audhkhasi2021mixture, lu2020exploring, chen2021developing, wu2020streaming, shi2021emformer, tsunoo2019transformer}. Though conceptually similar to block processing (e.g. \cite{tsunoo2019transformer, shi2021emformer}), chunk-wise attention is more flexible.  Block processing is performed at the input feature level, which limits the encoder layers to the context frame at the current chunk.  On the other hand, chunk-wise attention allows other layers in the encoder (e.g., convolution layers) to process contextual frames beyond the current chunk. Compared with Whisper \cite{radford2022robust}, which segments the audio into 30 second chunks and uses a heuristic process to carry the decoder states over, we only chunk the attention state, and allow the decoder to access the entire encoder output. We also use either a CTC or RNN-T decoder to decode on long-form audio, neither of which have been observed to hallucinate compared to attention-based sequence-to-sequence decoders. We observe our systems are robust on long-form ASR tasks with a simpler decoding process on long-form speech signals.

\subsection{Multi-Objective Supervised Pre-training: BEST-RQ + text-injection}
\begin{figure}[h!]
\centering
\includegraphics[width=0.9\columnwidth]{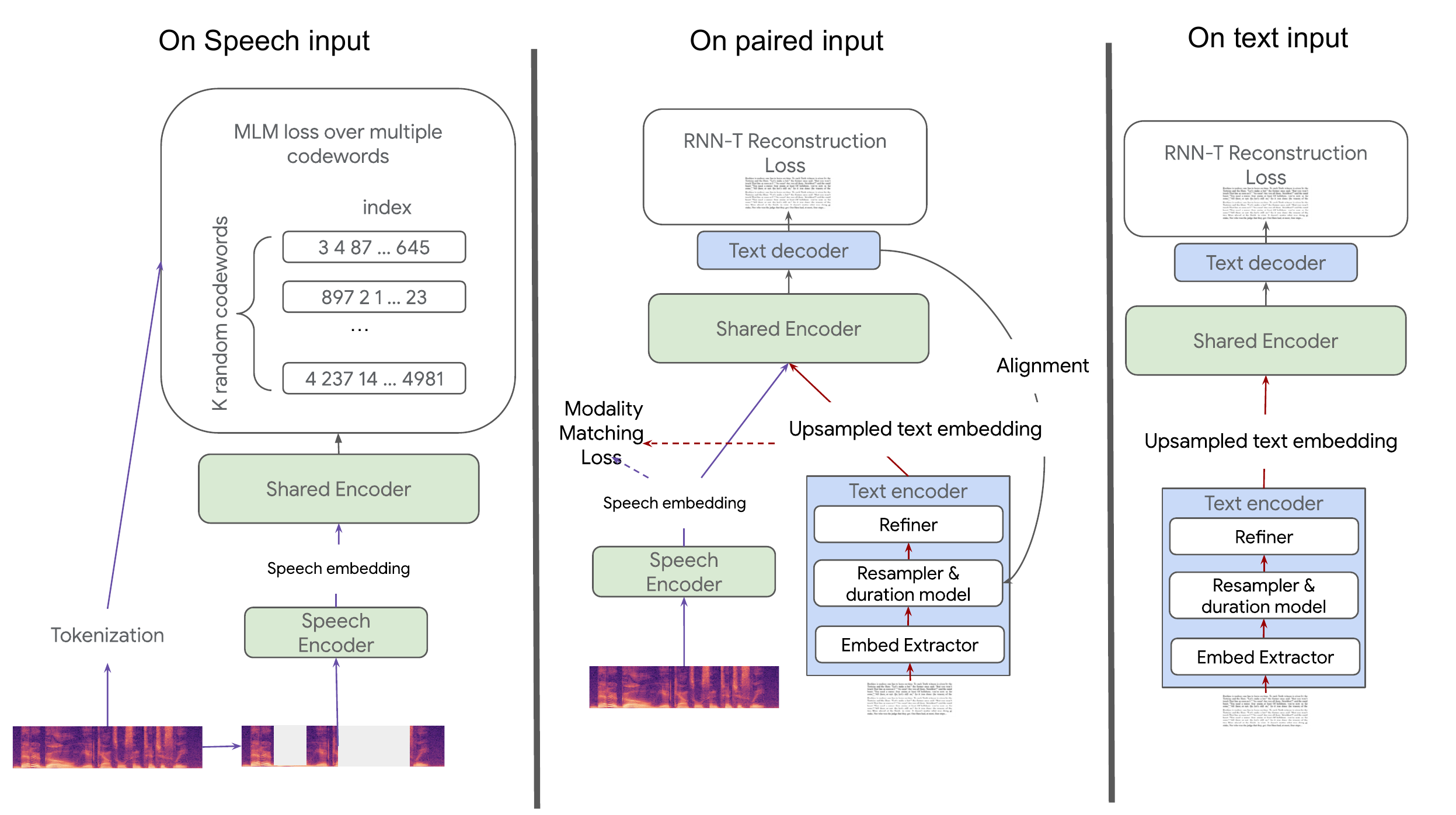}
\caption{Overview of MOST text injection. The left-most panel depicts MOST training on unlabeled speech input; the center panel depicts training on paired speech and text input; the right-most panel depicts training on unlabeled text data.}
\label{f:text-injection}
\end{figure}
In addition to pre-training with unlabeled speech, we add an additional stage of \textbf{M}ulti-\textbf{O}bjective \textbf{S}upervised pre-\textbf{T}raining (MOST) as shown in Fig. \ref{f:text-injection}, where we train the model jointly on unlabeled speech, unlabeled text and paired speech and text data. The training loss for this procedure is based on the text-injection loss including duration modeling and consistency regularization as in \cite{chen2022maestro}, to which we add a weighted BEST-RQ loss for the encoder of the model. MOST yields two benefits: (i) Training with paired speech and text data with alignment losses results in learning speech representations that are better aligned with text, improving quality on tasks like ASR and AST that require mapping the acoustics of the speech signal to text. (ii) Training simultaneously on unlabeled text in a model that learns speech and text representations jointly improves the robustness of learned representations, especially on low resource languages and domains, also generalizing to new languages with no paired data seen during training~\cite{chen2022maestroU}.

The key architectural components for constructing the text-injection loss as utilized in our approach include: (i) A speech-only encoder that utilizes a convolutional sub-sampling feature encoder and a single conformer layer. For continued pre-training the feature encoder is initialized from the BEST-RQ pre-trained checkpoint while the conformer layer is initialized randomly. (ii) A text-only encoder that consists of an embedding layer, an upsampler, and a conformer layer block. The upsampler used in this work is a learned duration based upsampling model \cite{chen2022maestro}, though a fixed or random repetition upsampler can also be used for text-injection \cite{thomas2022textogram,sainath2023joist}. All components are initialized randomly. (iii) A shared conformer encoder initialized from the pre-trained BEST-RQ speech encoder. (iv) The BEST-RQ speech softmax layers initialized from the BEST-RQ checkpoint. (v) The decoder unit which is initialized randomly.

The main idea of text-injection (e.g.~ \cite{chen2022maestro,sainath2023joist,zhong2023jeit}) is to produce joint, co-aligned embeddings of speech and text as sequences in the same embedding space. Given this embedding space, text data with no associated audio can contribute to improving the speech task. The speech and text encoders presented above are intended to produce these embeddings, which need to be matched in the embedding space and are also required to be co-aligned in the time dimension. The embeddings enable the text data to contribute to preparing the model for downstream tasks.

To achieve these objectives, the architecture as presented above is trained using three types of data, each contributing to different types of losses:
\begin{enumerate}
\item The unlabeled speech passes through the shared encoder and the BEST-RQ softmax layers to contribute to the BEST-RQ loss.
\item The paired speech-text data serves multiple functions.
 \begin{itemize}
 \item The labeled speech flows through the speech encoder, the shared encoder and the decoder unit and contributes to the standard ASR loss computed against the paired text. Here, the speech-text alignments of the paired data are extracted from the decoder unit and used to train the duration upsampler within the text encoder.
 \item The text of the paired data also passes through the text encoder. The encoded text sequence is used to compute a consistency loss against the encoded speech sequence. This loss is used to train solely the text encoder---the speech encoder weights are frozen for this particular forward-propagation.
 \end{itemize}
\item The unlabeled text data contributes to a reconstruction loss. This loss is constructed by passing the text through the text encoder, then masking chunks of the feature sequence produced. These masked text features live in the same embedding space as masked speech features, and thus can be passed through the shared encoder and the decoder unit to compute the ASR loss against the original text. This is the reconstruction loss used to train the model.
\end{enumerate}

For training stability, MOST proceeds in two stages---we first train solely on paired data to learn stable decoder alignments for 20k steps. We then train the duration upsampler and activate the losses for unlabeled text. We refer the reader to \cite{chen2022maestro} for further details.

When fine-tuning for ASR, we initialize the feature encoder of the ASR model with the speech feature encoder, initialize the conformer block with the shared conformer encoder, and add a randomly initialized task-specific transducer. 

In the MOST set-up, the speech and text representations live in a shared representation space, thereby allowing us to utilize text machine translation (MT) data during the fine-tuning stage of AST tasks. We follow the same approach described in ~\cite{bapna2022mslam,chen2022maestro} and report the AST results with joint fine-tuning for models prepared with MOST.

\subsection{Residual Adaptation with a Frozen Encoder}

Ideally, the fine-tuning process of the model should be scalable with the number of downstream tasks while in reality, fine-tuning the pre-trained USM individually for various domains and tasks becomes prohibitively expensive. In order to mitigate this issue, we explore a lightweight alternative \cite{he2022towards} to training the full network where residual adapters with a small number of parameters are added for each individual language while the pre-trained USM is entirely frozen during fine-tuning. We experiment with adding two parallel adapters to each Conformer block, whose parameter count amounts to 2\% of the original pre-trained USM, and fine-tune the adapters on downstream language tasks. When serving the model, the adapter is dynamically loaded according to the language of the input batch~\cite{BiadsyCZRRM22, tomanek2021residual}. This enables one to conduct inference on 100+ languages while keeping the total number of parameters manageable by re-using the same parameters and computation process for the majority of the time. We also find that training the adapter versus fine-tuning the entire model can reduce over-fitting especially when the training data is limited.

\subsection{Training Details}
\label{ss:details}

\textbf{Data Processing: } The audio is uniformly sampled to 16 kHz quality---any audio with a different native sampling rate is either up-sampled or down-sampled. The audio is then featurized into 128-dimensional log-mel filterbank coefficients. Graphemes are used to tokenize the text for FLEURS in-domain fine-tuning, while word-piece models (WPMs) \cite{wpm} are used for tokenization for all other tasks.

\textbf{BEST-RQ: } We follow default masking and quantization parameters of BEST-RQ as in \cite{pmlr-v162-chiu22a}. We use a 16 codebook multi-softmax loss to stabilize training and improve performance as described in \ref{sec:multisoftmax}. We do not use EMA for pre-training.

\textbf{MOST: }
We follow the text encoder and decoder architecture described in \cite{chen2022maestro} but use 4k sentence-piece models (SPMs). We use a single 1536-dimensional Conformer layer as the speech encoder and Conformer-2B encoder as the shared encoder. We mix un-transcribed speech, unspoken text, and transcribed speech in each batch with fixed batch sizes of, respectively, 4096, 8192, and 1024. The model is initialized with the BEST-RQ pre-trained encoder. MOST employs a curriculum learning schedule where training initially is conducted with un-transcribed speech and paired speech-text data, and unspoken text is utilized only after 20k steps. The joint training employing all three types of data lasts for another 100K steps.

\textbf{Supervised Training: } We use two separate optimizers for the encoder parameters and the decoder parameters of the network \cite{ssllimit}. For USM-CTC and USM-LAS, we train the model for 100k steps with 2048 batch size. For in-domain experiments, the checkpoint is selected based on development set performance.

\textbf{Training Large Models: } We use the GShard~\cite{gshard-arxive} framework with the GSPMD backend~\cite{gspmd} to train our large models on TPUs.

\section{Datasets}
\label{sec:data}
\subsection{Audio Data}
\label{ss:ytdata}

\begin{figure}[h!]
\centering
\includegraphics[width=0.98\columnwidth]{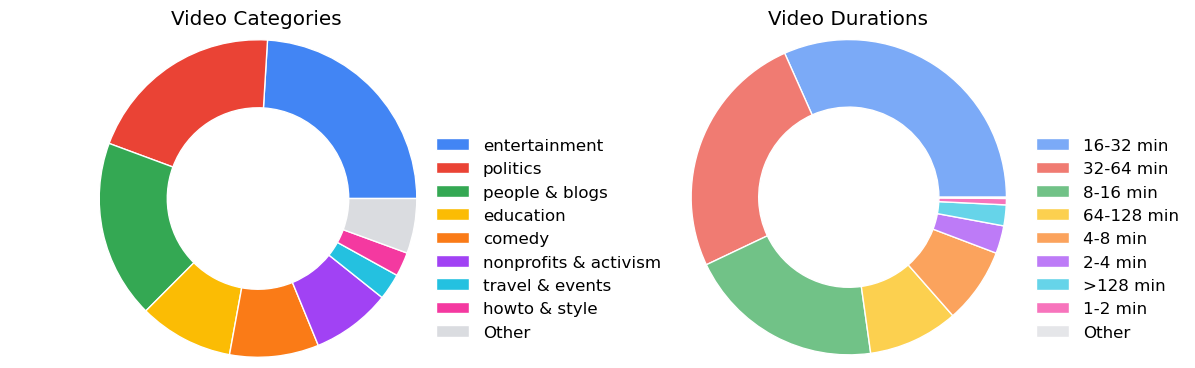}
\caption{The video category and length distribution of YT-513-U.}
\label{f:yt513_video_distribution}
\end{figure}

The following audio datasets are used in this report to train our models:
\begin{itemize}
\item \textbf{YouTube SUPervised Plus (YT-SUP+)}: 
    \begin{itemize}
        \item YT-SUP: 90k hours of segmented, labeled audio across 75 languages.
        \item YT-Pseudo-Labeled: 100k hours of segmented, pseudo-labeled en-US audio from YT-NTL-U. The pseudo-labels are generated by a 600M CTC model trained on YT-SUP en-US data.
    \end{itemize}
\item \textbf{YouTube Next Thousand Languages Unsupervised (YT-NTL-U)}: 12.1M hours of segmented, unlabeled audio, including:
\begin{itemize}
\item YT-55-U: 12M hours of segmented, unlabeled audio on 55 rich resource languages identified by YouTube production language id models.
\item YT-513-U: 100k hours of segmented, unlabeled audio across 513 tail languages not covered by YouTube production language id models. These languages are identified by vendors.
\end{itemize}
\end{itemize}

Let us expand upon how each dataset has been constructed.

\textbf{YT-SUP+:} YT-SUP is a dataset with audio from videos that have user-uploaded transcripts from 75 languages. We group consecutive segments into a longer unit similar to \cite{lu2021input}. The maximal sequence length for training is 30 seconds. The total amount of training data is 90k hours, ranging from English (en-US) (3.5k hours) to Amharic (Am-Et) (150 hours). We also introduce an additional 100k hours of en-US audio from YT-NTL-U to YT-SUP. We choose to generate pseudo-labels on this dataset using a 600M-parameter CTC YT teacher model trained on YT-SUP. Each audio is randomly segmented between 5 to 15 seconds.

\textbf{YT-55-U: } YT-55-U is built by first randomly collecting 3 million hours of audio from "speech-heavy" YouTube videos, filtered by language. The 3 million hours of audio is then further segmented by the YT teacher model. Instead of using a teacher model as in \cite{bigssl}, the non-speech segments identified by a Voice Activity Detection (VAD) model are removed to yield approximately 1 million hours of unlabeled audio data. Later, we use a YouTube production language identification model to select 55 languages from that audio.

\textbf{YT-513-U: }We create an additional dataset called YT-513-U to ensure coverage of lower resource languages in our pre-training dataset.
We reached out to vendors and native speakers to identify YT videos containing speech in specific long tail languages, collecting a dataset of unlabeled speech in 513 languages. Vendors were tasked with ensuring a variety of domains, voices, and content in the videos that are collected in each language. These videos are segmented into speech segments using a VAD model, resulting in a total of 102k hours of speech. Our final YT-513-U dataset contains 88 languages with over 500 hours of speech each, 237 languages with between 100-500 hours, and 188 languages with less than 100 hours of data.  The languages chosen for this collection are wide-ranging, with a majority of our data corresponding to languages from South Asia, Southeast Asia, West Africa, and East Africa. The distribution of video categories and lengths in our dataset are depicted in Figure~\ref{f:yt513_video_distribution}.

In addition to YouTube data, we also include public data for \textbf{MOST} training:
\begin{itemize}
    \item \textbf{Public Unsupervised (Pub-U)}: Following \cite{bapna2022mslam}, we use approximately 429k hours of unlabeled speech data in 51 languages. It includes: 372k hours of speech data spanning 23 languages from VoxPopuli~\cite{wang2021voxpopuli}, read speech data in 25 languages drawn from the v6.1 release of Common Voice~\cite{ardila2019common}, 50k hours of read books data in eight European languages from Multilingual LibriSpeech~\cite{pratap2020mls} and 1k hours of telephonic conversation data spanning $17$ African and Asian languages from BABEL~\cite{Gales2014SpeechRA}.
    \item \textbf{Public Supervised (Pub-S)}: Similar to \cite{bapna2022mslam}, our public supervised set includes approximately 1.3k hours of speech and transcript data spanning 14 languages from VoxPopuli, $10$ hour training splits for each of the $8$ MLS languages, and $1k$ hours of data spanning $17$ languages from the Babel ASR task.
\end{itemize}

Note that the public data is only used for in-domain pre-training and is excluded for training the generic USM-LAS/CTC models.  This allows us to treat the public task performance as out-of-domain benchmarks for the USM-LAS/CTC models.

\subsection{Text Data}

\textbf{Web-NTL: } For pre-training with unlabeled text, we use a web-crawled corpus of monolingual text containing over 28B sentences~\cite{bapna2022building}. The dataset spans $1140$ languages, $205$ of which have over $1M$ sentences and $199$ of which have between $100k$ and $1M$ sentences. We up-sample lower resource languages using temperature-based sampling~\cite{arivazhagan2019massively} with $T=3.0$. More details about the dataset and the mining approach have been described in Section 2 of~\cite{bapna2022building}. 

\subsection{Downstream Benchmarks}

\subsubsection{Speech Recognition (ASR)}
We present our results on two public tasks, SpeechStew \cite{speechstew} and FLEURS \cite{conneau2022fleurs}, and an internal benchmark on YouTube.

The \textbf{SpeechStew} \cite{speechstew} dataset is assembled by putting together seven public speech corpora---AMI \cite{ami}, Common Voice \cite{cv}, English Broadcast News%
\footnote{Linguistic data consortium (LDC) datasets LDC97S44, LDC97T22, LDC98S71 and LDC98T28.},
LibriSpeech \cite{librispeech}, Switchboard/Fisher%
\footnote{LDC datasets LDC2004T19, LDC2005T19, LDC2004S13, LDC2005S13 and LDC97S62.},
TED-LIUM v3 \cite{rousseau2012ted,hernandez2018ted} and Wall Street Journal\footnote{LDC datasets LDC93S6B and LDC94S13B.}, which are all standard benchmarks \cite{gauvain1994limsi,kubala19981997,chen1998ibm} covering different domains in en-US. 

The \textbf{FLEURS} \cite{conneau2022fleurs} dataset is a publicly available, multi-way parallel dataset of $10$ hours of read speech in $102$ languages spanning $7$ geo-groups. We restrict our use of the dataset to its ASR benchmark. Among the $102$ languages present in the FLEURS benchmark, we select $62$ to serve as a sub-group to compare our generic ASR system with Whisper \cite{radford2022robust}, as those languages are covered by the training sets of both models. We also report full results for in-domain fine-tuning and adaptation. Unlike \cite{conneau2022fleurs}, we report both WER and CER metrics, as CER is inappropriate as an indicator of performance for some languages. When presenting the error rate metrics, we use CER for Chinese, Japanese, Thai, Lao, and Burmese to be consistent with Whisper \cite{radford2022robust}.

The test set for the \textbf{YouTube} domain consists of utterances from 73 languages with an average of 15 hours of audio per language, the audio length for each individual language ranging from 1 to 24 hours. The audio is transcribed manually from popular YouTube videos, each with a duration of up to 30 minutes.

\subsubsection{Speech Translation (AST)}
Following \cite{bapna2022mslam}, we use CoVoST 2 \cite{wang2020covost} to benchmark multilingual speech translation. We evaluate the multilingual XX-to-English task that covers translation from 21 source languages into English. Depending on the language, the training data ranges in size from 1 - 264 hours.

Besides speech translation data, we also add text-to-text translation data for training the model as in \cite{bapna2022mslam}. This dataset includes the text translation data from CoVoST 2 combined with all data from either WMT or TED Talks, as available.
\section{Key Results}
\label{s:downstreamasr}

\begin{table*}[t]
\caption{WERs (\%) across multiple tasks for multiple settings compared against pre-existing baselines, with the exception of CoVoST 2, for which the BLEU score is presented. For the YouTube long-form set, we select the top-25 languages Whisper was trained on and exclude all languages for which Whisper produces > 40\% WER to reduce the noise introduced by LAS hallucination in the Whisper model. For FLEURS, we report both the WER and the CER for our models. $^\dagger$Results omitted for the Whisper-shortform model on the YouTube long-form dataset as the model has a high deletion problem on this set. ${^\ddagger}$The Whisper-shortform model uses segmented decoding to reduce its hallucination problem on CORAAL.  ${^\mathsection}$Our adapter setup adds about 2.3\% of the total parameters while keeping the encoder frozen from pre-training.}
\centering
\vskip 0.1in
\resizebox{0.98\textwidth}{!}{%
\begin{tabular}{lcccccccc}
\toprule
\bfseries Task & \multicolumn{4}{c}{Multilingual Long-form ASR} &  Multidomain en-US &  \multicolumn{2}{c}{Multilingual ASR} & AST \\
\midrule
\quad Dataset & \multicolumn{3}{c}{YouTube} & CORAAL & SpeechStew & \multicolumn{2}{c}{FLEURS} & CoVoST 2\\
\quad Langauges & en-US & 18 & 73 &  en-US & en-US & 62 & 102 & 21 \\
\midrule
\bfseries Prior Work (single model)\\
\quad Whisper-longform & 17.7 & 27.8 & - & 23.9 & 12.8 &  \\
\quad Whisper-shortform$^\dagger$ & - & - & - & 13.2${^\ddagger}$ & 11.5 &  36.6 & - & 29.1 \\
\bfseries Our Work (single model)\\
\quad USM-LAS & 14.4 & 19.0 & 29.8 & \textbf{11.2} & \textbf{10.5} & \textbf{12.5} & - & - \\ 
\quad USM-CTC & \textbf{13.7} & \textbf{18.7} & \textbf{26.7} & 12.1 & 10.8 &  15.5 & - & - \\
\midrule
\bfseries Prior Work (in-domain fine-tuning)\\
\quad BigSSL \cite{bigssl}& 14.8 & - & - & - & 7.5 & - & - & -\\
\quad Maestro \cite{chen2022maestroU} & & & & & 7.2 & & & 25.2 \\
\quad Maestro-U \cite{chen2022maestroU} & & & & & & & 26.0 (8.7) \\
\bfseries Our Work (in-domain fine-tuning)\\
\quad USM & 13.2 & - & - & - & 7.4 & {13.5} & {19.2 (6.9)} & 28.7\\
\quad USM-M & \textbf{12.5} & - & - & - & \textbf{7.0} & \textbf{11.8} & \textbf{17.4 (6.5)} & \textbf{30.7}\\
\bfseries Our Work (frozen encoder)\\
\quad USM-M-adapter${^\mathsection}$  & - & - & - & - & 7.5 & 12.4 & 17.6 (6.7) & 29.6\\
\bottomrule
\end{tabular}}
\label{t:summary}
\end{table*}

\subsection{Robust Speech Recognition for Massively Multilingual Tasks}
In this section, we compare the performance of our models against public baselines, including Whisper large-v2\footnote{Whisper large-v2 on Github (https://github.com/openai/whisper.git,  revision b4308c4) is used for evaluation.}  \cite{radford2022robust}, which has been trained on 680k hours of weakly supervised data across 100 languages.

For the massively multilingual speech recognition test dataset from YouTube, we observe that Whisper hallucinates in many languages, resulting in a WER exceeding 100\%. For a reasonable comparison, we restrict the language set on which we compare the performance USM against Whisper by first selecting the top-25 languages from the training data for Whisper and further excluding languages for which Whisper produces > 40\% WER. We also use segmented decoding for Whisper with 30-second segments to further reduce the effect of hallucinations.
As shown in Table \ref{t:summary}, our USM-LAS and USM-CTC models outperform Whisper by a wide margin on YouTube en-US, despite training on significantly less supervised data (3.5k hours versus Whisper's 400k hours \cite{radford2022robust}). While the USM-LAS model also requires segmented decoding to reduce long-form degradation as discussed in section \ref{ss:chunk-wise-attention}, it is far more robust, out-performing Whisper by a relative 30\% WER on those 18 languages. USM-CTC does not exhibit long-form performance degradation and achieves the best performance on YouTube. 

On the out-of-domain long-form CORAAL set, both USM-CTC and USM-LAS outperform Whisper by more than 10\% relative WER. USM-CTC and USM-LAS similary outperform Whisper on SpeechStew, whose training data the models have not had access to.

We further compare the multilingual performance of the models on the held-out set from FLEURS. As shown in Table \ref{t:summary}, USM-LAS and USM-CTC both outperform Whisper by 66\% relative WER, despite using a smaller amount of multilingual supervised data (90k versus Whisper's 117k, when en-US is excluded). USM-LAS consistently outperforms USM-CTC for short-form ASR tasks.

\subsection{Massively Multilingual Results Beyond 100 Languages}

The lower part of Table \ref{t:summary} shows our results for in-domain fine-tuning. Our pre-trained model improves the FLEURS benchmark significantly, even when using only 10 hours per language. Compared to the previous SoTA in \cite{chen2022maestroU}, our model achieves a 30\% relative improvement in terms of WER across 102 languages. Our results show that while generic speech models can be powerful, performance is still maximized by in-domain fine-tuning.

\subsection{MOST Produces Robust Representations that Generalize to New Domains}

MOST training aligns the representations of speech and text by training simultaneously on the two modalities. We investigate whether MOST representations are useful for adapting the model to new domains by freezing the entire learned encoder produced by MOST and adjusting a small amount of parameters added to the network by residual adapters.  As shown in Table \ref{t:summary}, by adding only 2\% to the total number of parameters, the MOST representation model (USM-M-adapter) only performs slightly worse than the fine-tuning baselines, still showing competitive performance on downstream ASR and AST tasks. The small number of parameters being trained in this approach makes it feasible to extend our system to a large number of new domains and new tasks, even with a limited amount of training data, such as in FLEURS.

\subsection{Pushing the Quality of ASR on Unseen Languages}
\begin{table}[h!]
  \caption{Noisy student training for unseen languages. WERs (\%) for the teacher adapter models and the student models are presented. The relative improvement (\%) of the student models can be found in the last column.}
  \label{tab:nst}
  \centering
  \resizebox{\textwidth}{!}{\begin{tabular}{lrrrrr}
    \toprule
    Languages & Whisper-v2 & \# hrs in YT-NTL & USM-LAS-Adapter & USM-M + pseudo label & Rel. Imprv. \\
    \midrule
     Hausa (ha)     & 88.9       & 2175.0 & 24.5 & 22.8 & 7.5 \\
     Kazakh (kk)     & 37.7       & 196.0  & 11.8 & 10.9 & 8.3 \\
     Shona (sn)     & 121.0      & 247.0  & 29.1 & 22.2 & 31.1 \\
     Pashto (ps)     & 93.7       & 254.0  & 36.0 & 35.4 & 1.7 \\
     Yoruba (yo)     & 94.8       & 1292.0 & 33.4 & 30.6 & 9.2 \\
    \bottomrule
  \end{tabular}}
\end{table}

Tail languages often do not have paired transcriptions for supervised learning---we refer to these languages as unseen languages, as the model has not seen paired data for these lanugages during training. To create pseudo-labels for these languages, we first build a USM-LAS-Adapter by attaching residual adapters to USM-LAS and training them using FLEURS data. By using the USM-LAS-Adapter as a teacher, we can now transcribe the unlabeled data in the unseen languages as part of the YT-NTL dataset.  As shown in Table 4, we observe consistent wins for all languages on the FLEURS benchmark.  For some languages, the improvement is larger than 30\%.  This further demonstrates the robustness of the USM-LAS model---despite using only 10 hours of out of domain data from FLEURS, the USM-LAS-Adapter is able to transcribe YouTube data to produce meaningful recognition results that lead to these improvements. We find the approach of training adapter models on small datasets and utilizing them for pseudo-labeling to be a promising route for scaling up the number of languages that can be transcribed by USMs.

\subsection{USMs are Strong AST Models}

The multi-lingual speech translation performance of fine-tuned USMs are shown in Table 3.  We find that we are already comparable to the CoVoST 2 SoTA BLEU score by fine-tuning the speech-only USM. We note that the previous SoTA uses 125k hours of supervised speech translation data compared to the 859 hours of data used by the USM. After MOST training, USM-M can use both speech and text as training input. By introducing text-to-text machine translation (MT) data during fine-tuning, USM-M is able to achieve an unprecedented > 30 BLEU on CoVoST (a 1 BLEU increase from SoTA). 

\section{Analysis and Ablations}
\label{sec:ablation}
\subsection{Multi-Softmax Loss for BEST-RQ}
\label{sec:multisoftmax}

We observe a consistent > 5\% relative improvement in ASR and AST benchmarks by increasing the number of the softmax groups in the multi-softmax loss for BEST-RQ training from 1 to 16, as shown in Table \ref{tab:language_coverage}. We also find that using multiple softmax groups significantly reduces performance variation across different pre-training runs and improves convergence speed.

\begin{table}[h!]
  \caption{YT-55 versus YT-NTL across different domains, with and without multi-softmax groups. For simplicity, we report CER for FLEURS. For CoVoST, we report the BLEU score. YT-NTL covers 27 additional languages not covered in YT-55.}
  \label{tab:language_coverage}
  \centering
  \resizebox{0.95\columnwidth}{!}{\begin{tabular}{lcccccc}
    \toprule
    Model & pre-train Set & \# Params (B) & \# Softmax & \multicolumn{2}{c}{FLEURS (CER)} & CoVoST (BLEU) \\
          &.             &        &       & 102 langs & 27 langs & \\
    \midrule
    Conformer-0.6B & YT-55 & 0.6 & 1 & 9.5 & - & 20.9 \\
    Conformer-2B & YT-55 & 2.0 & 1 & 7.9 & 9.5 & 26.6 \\
    Conformer-2B & YT-NTL-U & 2.0 & 1 & 7.4 & 8.5 & 27.5 \\
    Conformer-2B & YT-NTL-U & 2.0 & 16 & 6.9 & 8.1 & 28.7 \\
    \bottomrule
  \end{tabular}}
\end{table}

\subsection{Model and Language Scaling}

We find that scaling up the model size and increasing the language coverage of the pre-training dataset greatly benefits the performance of the USMs, as demonstrated in Table \ref{tab:language_coverage}. In particular, we find a 10\% relative improvement of ASR and AST performance by using YT-NTL vs. YT-55 for pre-training, despite the fact that each newly added language in YT-NTL contains approximately 500 hours of speech---a relatively small amount. As could be expected, the relative gains on the newly covered languages are more substantial than those on other languages.

\subsection{BEST-RQ is a Scalable Self-supervised Learner}
\label{sec:best-rq}

\begin{table}[hbt!]
\centering
\caption{BLEU scores for the CoVoST 2 X $\rightarrow$ En task to compare BEST-RQ against W2v-BERT. Higher is better.}
\label{tab:ablation_pretrain_covost}
\small\begin{tabular}{llrrrr}
\toprule
\xenglish{} && high & mid & low & all \\
\midrule
\textbf{Previous Work} \\[2pt]
~~~~\xlsrpb{0.3}~\cite{babu2021xls}  && 30.6 & 18.9 & 5.1 & 13.2 \\
~~~~\xlsrpb{1}~\cite{babu2021xls}  && 34.3 & 25.5 & 11.7 & 19.3 \\
~~~~\xlsrpb{2}~\cite{babu2021xls}  && 36.1 &   27.7 &   15.1 &   22.1 \\
\midrule
\textbf{Conformer-0.6B} \\[2pt]
~~~~W2v-BERT && 35.6 & 25.3 & 13.4 & 20.4 \\
~~~~BEST-RQ &&  32.5 & 25.6 & 14.7 & 20.7 \\
\midrule
\textbf{Conformer-2B} \\[2pt]
~~~~W2v-BERT && 36.0 & 27.8 & 15.6 & 22.4 \\
~~~~BEST-RQ && 35.8  & 31.3 & 21.5 & 26.6 \\
\bottomrule
\end{tabular} 
\end{table}

BEST-RQ has been shown to outperform or be comparable to other prominent pre-training methods for speech recognition, including wav2vec 2.0 and W2v-BERT in the original work in which it was introduced \cite{pmlr-v162-chiu22a}. Here we investigate its comparative performance and scaling properties, similar to what has been done for wav2vec 2.0 in \cite{bigssl} and W2v-BERT in \cite{bapna2022mslam}. We utilize the set-up of pre-training the model using YT-55 and fine-tuning it on CoVoST 2. As shown in Table~\ref{tab:ablation_pretrain_covost}, our results indicate that for the Conformer-0.6B, W2v-BERT and BEST-RQ perform similarly, but BEST-RQ obtains greater gains when scaled up. A contributing factor to this can be that W2v-BERT is more prone to codebook collapse and training instabilities at the 2B scale, while BEST-RQ by construction doesn't suffer from codebook collapse.

\subsection{Chunk-wise attention for robust long-form speech recognition}

\begin{figure}[h!]
\centering
\includegraphics[width=0.9\columnwidth]{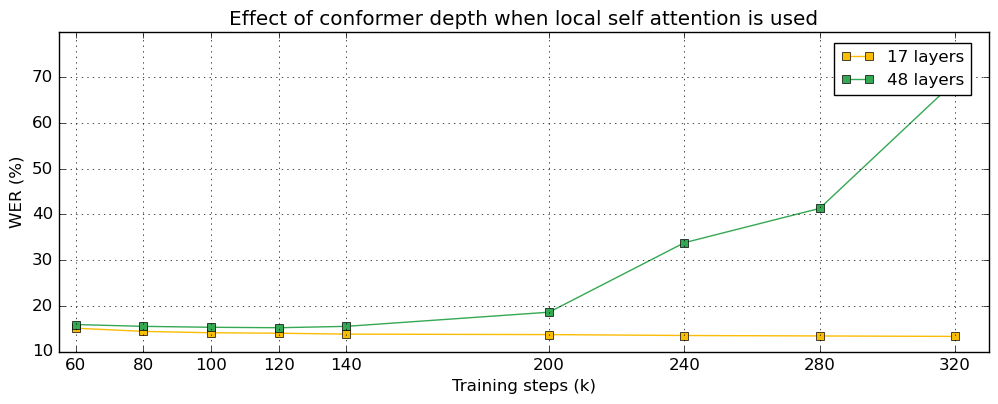}
\caption{The word error rate measured on the YouTube en-US long-form test set for Conformer models with varying depth.}
\label{f:conformer_depth_and_lsa}
\end{figure}

Fig. \ref{f:conformer_depth_and_lsa} depicts the long-form performance degradation issue as described in section \ref{ss:chunk-wise-attention}. In the figure, we see that for the shallow Conformer model with 17 layers, using a small local self attention context (65) length, the word error rate measured on the long-form test set gradually improves as the training progresses. With a deeper model that has 48 layers but roughly the same number of parameters, however, the larger receptive field mismatch results in higher test WERs as the training step increases.

Table \ref{tab:chunkwise} demonstrates that chunk-wise attention is able to address the long-form degradation issue and show robust performance across four different languages---en-US (English), ru-RU (Russian), ko-KR (Korean), and uk-UA (Ukrainian). We compare chunk-wise attention models with an 8-second chunk size (\texttt{CW-8s} in Table \ref{tab:chunkwise}) against local self attention models which uses 128 context frames in each conformer layer (\texttt{LSA-128}). We note that further increasing the context window size of the local self attention model results in high deletion error rates on all languages of the YouTube long-form test sets. These results show that the chunk-wise attention models do not exhibit long-form performance degradation and are able to improve upon the performance of the local self attention models operating at the maximum allowed receptive field length.
\begin{table}[h!]
  \caption{Chunk-wise attention. WER (\%) is reported on the YouTube long-form set.}
  \label{tab:chunkwise}
  \centering
  \begin{tabular}{lcccccccc}
    \toprule
    Model & \# Params (B) & \# Layers & en-US & ru-RU & ko-KR & uk-UA \\
    \midrule
    \texttt{LSA-128} & 0.6 & 24 & 16.2 & 16.6 & 26.2 & 15.5 \\
    \texttt{CW-8s} & 0.6 & 24 & 12.5 & 14.7 & 19.5 & 15.3\\
    \bottomrule
  \end{tabular}
\end{table}

\subsection{TPU Serving Capacity of USM-CTC Models}

\begin{table}[h!]
  \caption{RTF for USM-2B.}
  \label{tab:rtf}
  \centering
  \begin{tabular}{lcccccc}
    \toprule
    Model & bf-16 & Streaming & \# Params (B) & TPU \cite{jouppi2021ten} & Batch Size & 1.0/RTF \\
    \midrule
    Conformer-0.1B  & Y & Y & 0.1 & TPUv4i  & 64 & 3047 \\
    Conformer-0.6B & N & N & 0.6 & TPUv4i & 64 & 1920 \\
    Conformer-2B  & N & N & 2.0 & TPUv4i & 32 & 827 \\
    \bottomrule
  \end{tabular}
\end{table}

In section \ref{s:downstreamasr}, we have demonstrated that USM-CTC models are powerful generic ASR models with reliable long-form transcription performance and excellent generalization properties. Here we measure the serving capacity of the USM-CTC model as represented by the real time factor (RTF) in an ideal setup where we assume that each batch sent to TPU is fully packed along the time axis. The results of these measurements are presented in Table~\ref{tab:rtf}. Surprisingly, we find that the 2B-paramter USM-CTC model is only $3.9\times$ slower than the 100M-parameter streaming model \cite{li2021better}, primarily due to the fact that our models operate at batch processing mode. This result demonstrates that the USM-CTC can be used as an offline transcriber efficiently on TPUs (or GPUs).

\section{Discussion}

In this report, we put forward a practical and flexible approach for training speech understanding models capable of scaling speech recognition to hundreds of languages. We conclude the report with summarizing insights gained in the process:

\textbf{Unlabeled versus weakly labeled data:} We believe diverse unlabeled data is more practical to acquire for building usable ASR for tail languages than weakly labeled data. We have demonstrated that collaborating with native speakers to identify unsupervised data in hundreds of tail languages can be an effective route to improving recognition performance on low resource languages.

\textbf{In-domain data is best:} We have demonstrated that we can build a robust ASR system across many domains by utilizing a large amount of unsupervised data and a small amount of labeled data. Our results, however, also confirm that the most effective way to optimize the performance for a given domain is to use in-domain data to fine-tune the model.

\textbf{CTC vs RNN-T vs LAS:} The best transducer depends on the downstream task. A large pre-trained model with a frozen encoder can allow experimenters to test different transducers quickly and select the optimal transducer for their purpose.

\section*{Acknowledgments}
We would like to thank Alexis Conneau, Min Ma, Shikhar Bharadwaj, Sid Dalmia, Jiahui Yu, Jian Cheng, Paul Rubenstein, Ye Jia, Justin Snyder, Vincent Tsang, Yuanzhong Xu, Tao Wang, Anusha Ramesh, Calum Barnes, Salem Haykal for useful discussions. 

We appreciate valuable feedback and support from Eli Collins, Jeff Dean, Sissie Hsiao, Zoubin Ghahramani. Special thanks to Austin Tarango, Lara Tumeh, and Jason Porta for their guidance around responsible AI practices.

\small

\bibliographystyle{IEEEtran}
\bibliography{references}

\end{document}